# Approach to Predicting News ─ A Precise Multi-LSTM Network With BERT

**Chen, Chia-Lin**

**National Sun Yat-sen University** Computer Science and Engineering

m073040105@g-mail.nsysu.edu.tw

**Huang, Pei-Yu**

**The University of London** Management and Digital Innovation

amanda10702@gmail.com

**Yi-Ting Huang**

**Academia Sinica** Institute of Information Science

ythuang@iis.sinica.edu.tw

**Chun Lin**

**Academia Sinica** Institute of Information Science

clin@iis.sinica.edu.tw

# Abstract

Varieties of Democracy (V-Dem) [1] is a new approach to conceptualizing and measuring democracy and politics. It has information for 200 countries and is one of the biggest databases for political science. According to the V-Dem annual democracy report 2019, Taiwan is one of the two countries that got disseminated false information from foreign governments the most. It also shows that the ''made-up news'' has caused a great deal of confusion among Taiwanese society and serious impacts on global stability. Although there are several applications helping distinguish the false information, we found out that the pre-processing of categorizing the news is still done by human labor. However, unlike the machine, human labor may cause some mistakes and cannot work for a long time. The growing demands on automatic machines in the near decades show that while the machine can do as good as humans or even better, using machines can reduce humans' burden and cut down cost.

Therefore, in this work, we build a predictive model to classify the category of the news. The corpora we used contains 28358 news and 200 news which are scraped from the online newspaper **L**iberty **T**imes **N**et (LTN)[2] website and includes 8 categories: Technology, Entertainment, Fashion, Politics, Sports, International, Finance, and Health. At first, we use **B**idirectional **E**ncoder **R**epresentations from **T**ransformers (BERT)[3] for word embeddings which transform each Chinese character into a (1,768) vector. Then, we use a **L**ong **S**hort-**T**erm **M**emory (LSTM) layer to transform word embeddings into sentence embeddings and add another LSTM layer to transform them into document embeddings. Each document embedding is an input for the final predicting model, which contains two Dense layers and one Activation layer. And each document embedding is transformed into 1 vector with 8 real numbers, then the highest one will correspond to the 8 categories, and it is the predicted category of the news.

The implementation of our network can be found on GitHub (https://github.com/LanaChen0/Predict_News).

Keywords: BERT, LSTM, word embedding, Dense, Activation

---

[1] https://www.v-dem.net/en/
[2] https://www.ltn.com.tw/
[3] https://github.com/google-research/bert

# 1. Introduction

With the rapid development of technology, communication applications like Line and WhatsApp are continuously influencing the way humans transmit their information. Line is a popular application in Taiwan with 21 million active users monthly[4]. It is an application that enables users to send texts, images, videos and make calls online instantly. In February 2015, Line announced the 600 million users mark had been passed and 700 million were expected by the end of the year[5]. Taiwanese people tend to use this application for any transmitting information. That is to say, it is a hotbed for fake news created by people with bad intentions or ignorance. This kind of platform enables falsehood to be diffused significantly farther, faster, deeper, and more broadly than the truth in all categories (Politics, Health, Entertainment, etcetera) of information.

In political wise, Taiwan, a country full of democratic prosperity, which has a special election culture that focuses significantly on "promotions", for instance, there are several promotional banners and flags typically displayed on every election. Most Taiwanese people are therefore highly exposed to the political atmosphere and are willing to share their own opinions of each candidate to others not only in public but also on social media. The island's 23 million citizens were bombarded with disinformation and misleading content through Line, Facebook, and online chat groups, promoted by social media trolls with bad intention. According to Washington Post on December 18th, 2018[6], China's government is interfering with Taiwan's elections in the few past years, carrying out massive propagandas and social media campaigns that spread false news designed to influence Taiwanese. There are only a few official accounts in Line are doing fact-checking for the users, for example, Cofacts(真的假的)[7] and Mei-Yu-E(美玉姨)[8], which are chatbots that clarify false news. However, these chatbots still need human labor to categorize the news and determine the fake news. Therefore, as fake news is a globally growing issue and especially directly interfering with the Taiwanese political process, this is an imperative issue to deal with.

Our report is focusing on the pre-processing of fake news, which is news categorized. We found out that it is an essential step in predicting fake news to categorize news in advance. The corpora we relied on are the 8 categories of news from **L**iberty **T**imes **N**et(**LTN**) and using **BERT** for word embeddings and two LSTM layers for predicting news categories.

# 2. Model and Experiment

## 2.1 Model Explanation

We used **BERT** for transforming word embedding, and restricted 30 lines (subsequences) and 20 words for each line in a document. **BERT** is a powerful utility for transforming words to word embeddings supported by Google AI, and word embedding is a set which vocabulary are mapped to vectors of real numbers. The first LSTM layer transforms word embeddings into sentence embeddings and the second LSTM layer transforms sentence embeddings into a document embedding, following with two Dense layers for scaling. In the end, a Softmax layer is for calculating the answer (Choose the highest one among the eight categories).

---

[4] https://expandedramblings.com/index.php/line-statistics/
[5] https://en.wikipedia.org/wiki/Line_(software)
[6] https://www.washingtonpost.com/opinions/2018/12/18/chinas-interference-elections-succeeded-taiwan/
[7] https://tictec.mysociety.org/2018/presentation/cofact-factchecking
[8] https://www.businesstoday.com.tw/article/category/154769/post/201901020016/80%E5%BE%8C%E5%A5%B3%E7%94%9F%20%20%E4%B8%80%E6%89%8B%E5%82%AC%E7%94%9FLINE%E6%89%93%E5%81%87%E3%80%8C%E7%BE%8E%E7%8E%89%E5%A7%A8%E3%80%8D

For example, each document embedding was transformed into 1 vector with 8 real numbers, such as (8.74730467e-05, 1.11537855e-04, 1.27863220e-03, 9.55423748e-04,
5.19969326e-04, 9.96643186e-01, 5.63649974e-06, 3.98051459e-04), then the highest one among them will correspond to the answer, which is the predicted category of the news. Figure 1 and 2 are the hierarchy of multi-LSTM Network with BERT.

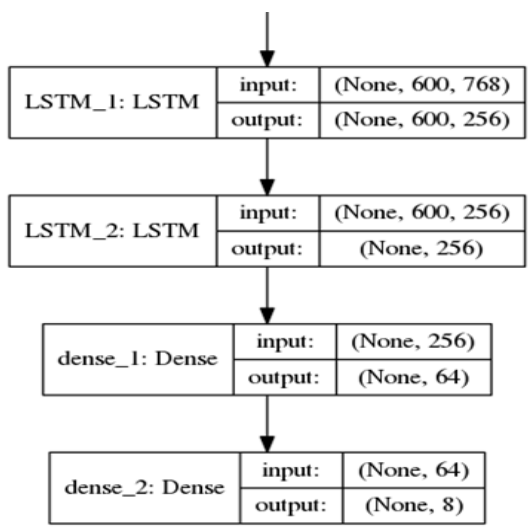

Figure 1.

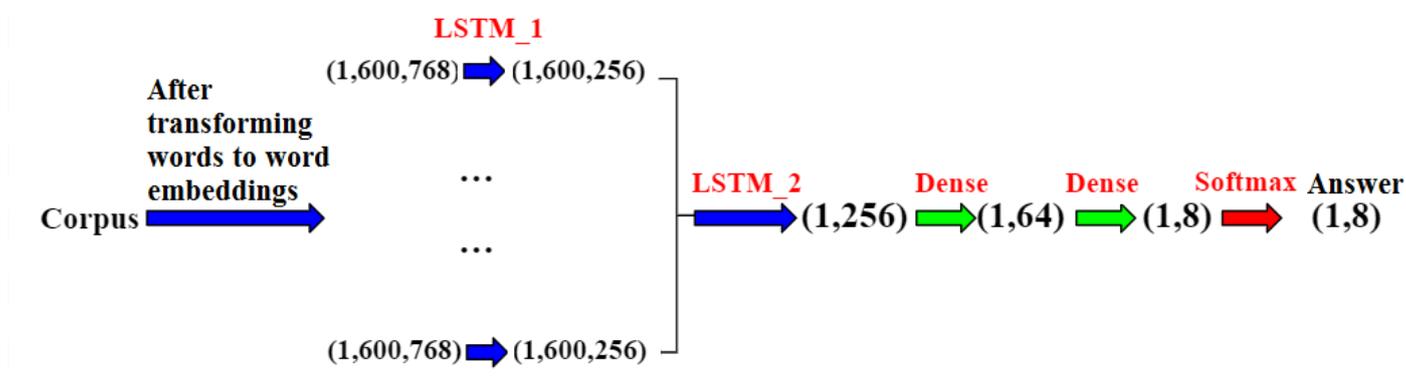

Figure 2.

## 2.2 Long Short-Term Memory(LSTM)

LSTM is commonly used in Neuro-Linguistic Programming (NLP[9]). LSTM is invented to adapt the main difficulty that Recurrent Neural Network (RNN[10]) has met, and it is immensely powerful in sequence prediction problems because they are able to store past information. LSTM has the architecture with four gates (input modulation gate, input gate, forget gate and output gate), and forget gate connects with two activation functions, tanh, and sigmoid functions, which in charge of seizing which information is necessary to memorize. Therefore, using LSTMs is a promising solution to sequence and time series related problems.

There are many utilities to implement LSTM, the library we use is Keras[11] with Tensorflow[12] as backend.

---

[9] https://en.wikipedia.org/wiki/NLP
[10] https://en.wikipedia.org/wiki/Recurrent_neural_network
[11] https://keras.io/
[12] https://www.tensorflow.org/

## 2.3 Hidden Layers in Transforming

There are two hidden layers in our two LSTM layers. In the first LSTM layer, inputs are 30 groups of 20-word embeddings and outputs are 30 sentence embeddings, it is a many-to-many process. However, in the second LSTM layer, 30 sentence embeddings, the inputs of this layer are also the outputs of the last layer, and the output is 1 document embedding which comes from the last hidden cell, and it is a many-to-one process.

## 2.4 Model Description

**d:** documents, **S:** sentences, **W:** words, **E:** word embeddings
**H:** hidden layers, **V**: concatenate of hidden layers, **D:** LSTM($V_0...V_{29}$)

| Layer | Description | Input | Output |
|---|---|---|---|
| 1 | Word to word embeddings : <br> E = BERT(W) | $W_0, …, W_{19}$ <br> $W_{20}, …, W_{39}$ <br> … <br> … <br> … <br> $W_{580}, …, W_{599}$ | $E_0, …, E_{19}$ <br> $E_{20}, …, E_{39}$ <br> … <br> … <br> … <br> $E_{580}, …, E_{599}$ |
| 2 | Word embeddings to sentence embeddings : <br> $H_0, …, H_{19}$ = LSTM($E_0, …, E_{19}$) <br> $V_0$ = concat($H_0, …, H_{19}$) <br> … <br> $V_k$ = concat($H_{20k}, …, H_{20k+19}$) <br> … <br> $V_{29}$ = concat($H_{580}, …, H_{599}$) | $E_0, …, E_{19}$ <br> $E_{20}, …, E_{39}$ <br> … <br> … <br> … <br> $E_{580}, …, E_{599}$ | $V_0$ <br> $V_1$ <br> … <br> … <br> … <br> $V_{29}$ |
| 3 | Sentence embeddings to a document embedding : <br> D = LSTM($V_0, …, V_{29}$) | $V_0, …, V_{29}$ | D |
| 4 | Document embedding to category label : <br> y = softmax(dense(dense(D))) | D | y (answer) |

## 2.5 Experiment

The following experiments include three versions of our model (version 0, 1 and 2), each with 200 and 28358 corpora sizes, 1 or 10 for batch sizes, and 5 types of document embeddings. The corpus with 200 news includes 2 Technology news, 78 Entertainment news, 52 Fashion news, 53 Politics news, 10 Finance news, and 5 Health news. And the corpus with 28358 news includes 6888 Technology news, 2265 Entertainment news, 5012 Fashion news, 2495 Politics news, 1974 Sports news, 1368 International news, 2324 Finance news, and 6032 Health news.

The first version (ver_0) is with only one BiLSTM layer and did word embedding without BERT; The second version (ver_1) is using word embedding with BERT, padding before BERT, and with two LSTM layers; The third version (ver_2) is using word embedding with BERT, padding after BERT, and with two LSTM layers. In our experiment, ver_2 is always performs better than others.

The selection of document embedding is always related to the input data. In our experiment, it was found out that a document with 600 words has always performed better than the others. For this reason, we chose 30 lines for each article and 20 words for each line, i.e. a document embedding size is (600,768).

Document Embedding (DE), DE_1 has only 1 line in each document and only 1 word in each line, i.e. a document embedding size is (1,768). DE_150 has 15 lines in each document and 10 words in a line, i.e. a document embedding size is (150,768). DE_1000_A has 100 lines in each document and 10 words in a line, i.e. a document embedding size is (1000,768). DE_1000_B is 10 lines in each document and 100 words in a line, i.e. a document embedding size is (1000,768).

All in all, in our experiments, models performed better and with more precise results when:

1. Using BERT for transforming words to word embeddings
2. Padding after doing BERT
3. Corpus with a larger size
4. With the smaller batch size

Table 1. Models Comparison

| Version | Corpora size | valid split | epoch | batch size | accuracy | val_acc | loss | val_loss |
|---|---|---|---|---|---|---|---|---|
| **ver_0 (DE_600)** | 200 | 8:2 | 10 | 10 | 41.25% | 40.00% | 1.2158 | 2.0627 |
| **ver_0 (DE_600)** | 200 | 8:2 | 10 | 1 | 63.12% | 4.54% | 0.9054 | 0.0250 |
| **ver_0 (DE_600)** | 28358 | 8:2 | 10 | 10 | 58.68% | 6.06% | 1.1852 | 3.8084 |
| **ver_1 (DE_600)** | 200 | 8:2 | 10 | 10 | 32.50% | 0% | 10.6782 | 16.1181 |
| **ver_1 (DE_600)** | 200 | 8:2 | 10 | 1 | 33.12% | 62.50% | 10.5775 | 2.3959 |
| **ver_1 (DE_600)** | 28358 | 8:2 | 10 | 10 | 99.06% | 99.48% | 0.0369 | 0.0230 |
| **ver_2 (DE_600)** | 200 | 8:2 | 10 | 10 | 77.91% | 4.88% | 0.6021 | 5.3912 |
| **ver_2 (DE_600)** | 200 | 8:2 | 10 | 1 | 88.96% | 43.90% | 0.4145 | 5.7418 |
| **ver_2 (DE_600)** | 28358 | 8:2 | 10 | 10 | 100.00% | 100.00% | 1.1921e-07 | 1.1921e-07 |
| **ver_2 (DE_1)** | 28358 | 8:2 | 10 | 10 | 35.71% | 58.71% | 10.3624 | 6.6550 |
| **ver_2 (DE_150)** | 28358 | 8:2 | 10 | 10 | 0.00% | 100.00% | 16.1181 | 1.1921e-07 |
| **ver_2 (DE_1000_A)** | 28358 | 8:2 | 10 | 10 | 100.00% | 0.00% | 8.1529e-06 | 12.4071 |
| **ver_2 (DE_1000_B)** | 28358 | 8:2 | 10 | 10 | 0.00% | 100.00% | 16.1181 | 4.2319e-06 |

ver_0: without BERT and only one BiLSTM layer (document to document embedding);
ver_1: padding before using BERT and two LSTM layers;
ver_2: padding after using BERT and two LSTM layers;
DE_1 is only 1 line in each news and only 1word in a line, i.e.(1,768).
DE_150 is 15 lines in each news and 10 words in a line, i.e. (150,768).
DE_600 is 30 lines in each news and 20 words in a line, i.e.(600,768).
DE_1000_A is 100 lines in each news and 10 words in a line, i.e.(1000,768).
DE_1000_B is 10 lines in each news and 100 words in a line, i.e.(1000,768).

Because ver_2 always performs better than ver_0 and ver_1 (according to Table 1), we printed out the line charts (Figure 3-5) of ver_2, which showed the differences when the features (embedding sizes, corpora size and batch sizes) have changed.

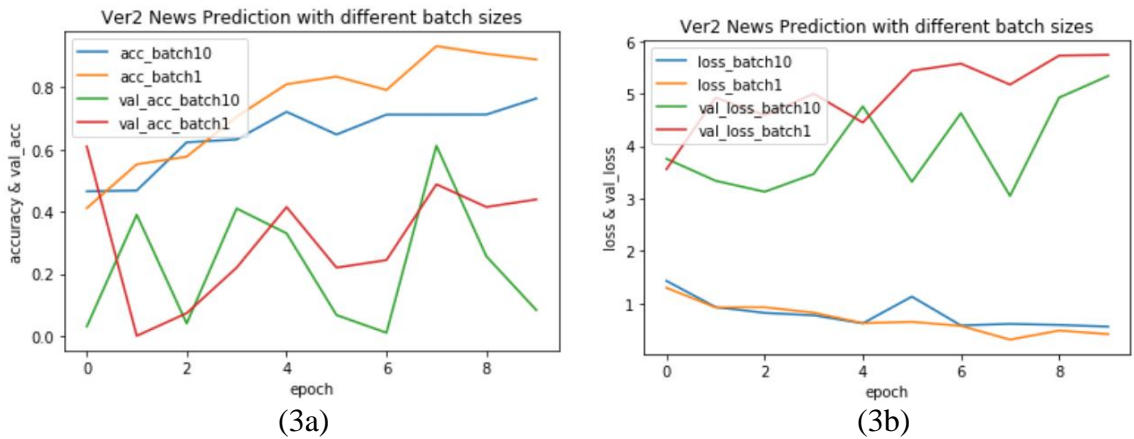

(3a)　　　　　　　　　　　　　　　　(3b)

Figure 3. Comparison of different batch sizes. Figure(3a)(3b) are all with same document embedding (600,768), valid split (8:2), epoch (10), corpora size (200) , but with different batch size 1, 10 in a document.
Batch size affects how many training data will be trained for each epoch. When the batch size becomes smaller, then the training size becomes larger, it needs more training time, and it also affects the precision. In our experiment, it was found that batch size = 1 was mostly better than batch size=10, and batch size=1 always did better in the tenth epoch (orange line can be compared with the blue line, red line can be compared with the green line) (also can be found in Table 1).
*Notice that the lines were not stable maybe because of less epochs (10 epochs for corpora size=200 was not enough).

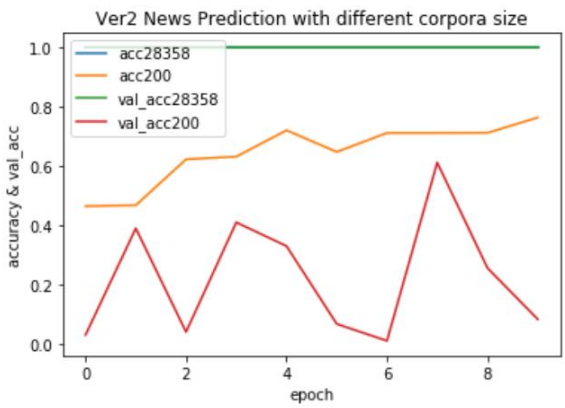 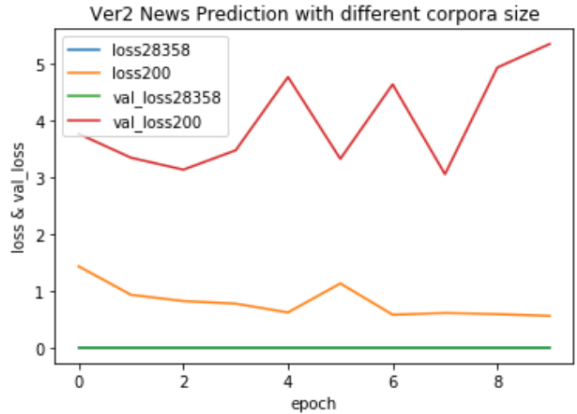

(4a)            (4b)

Figure 4. Comparison of different corpora sizes. Figure(4a)(4b) are all with the same document embedding (600,768), valid split (8:2), epoch (10), and batch size (10), but with different corpora size 28358, 200 in a document. Larger corpora size means more data were used for training, validating, and testing. In our experiment, corpora size=28358 is 100 times more than corpora size=200, so the model can get respectively accurate results.
*Notice that blue line and green line overlapped completely.
*Notice that the lines were not stable maybe because of less epochs (10 epochs for corpora size=200 was not enough).

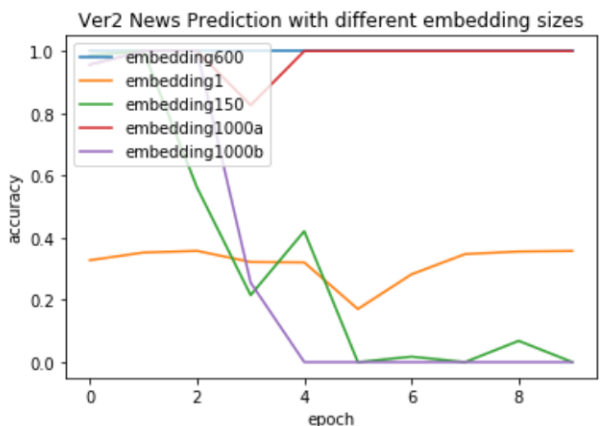 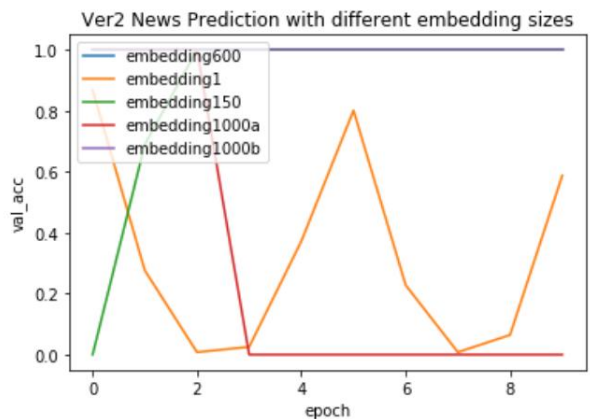

(5a)            (5b)

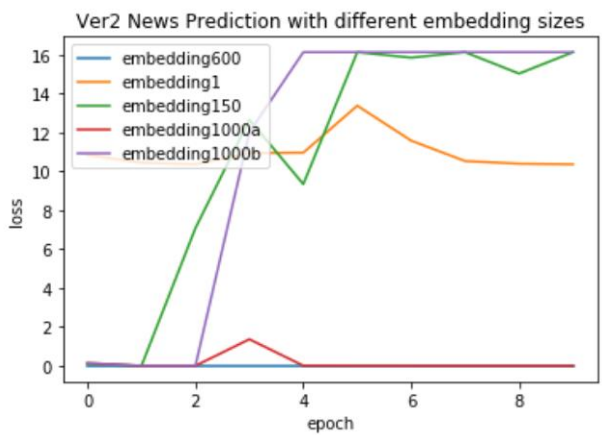 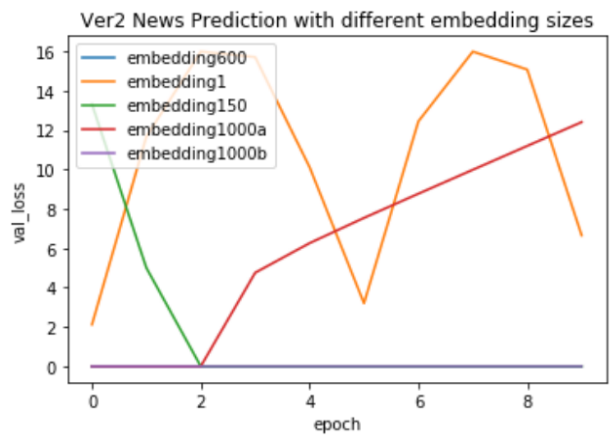

(5c)            (5d)

Figure 5. Comparison of different document embedding types using ver_2 model. Figure (5a)(5b)(5c)(5d) are all with the same corpora size (28358), valid split (8:2), epoch (10), batch size (10), but with different types of document embedding. We choose 1, 150, 600, 1000a, 1000b as document embeddings in each document, according to these pictures, it can be understood that only embedding600 worked well on both accuracy and loss test, although embedding1000b worked as good as embedding600 on accuracy, it did bad on validation loss test.

*Notice that blue line and red line overlapped partially in figure(5a)(5c), and blue line and purple line overlapped completely in figure(5b)(5d).

*Notice that some lines were not stable maybe because of less epochs (10 epochs for some types of document embedding was not enough).

One of the reasons caused some unstable lines maybe because of testing 10 epochs in the experiments, but this statement remains unsure. Only when we do more experiments, then we can be sure whether less epochs leads to stable lines or the lines are never stable.

## 3. Conclusion

There are similar experiments that focusing on classification on insincere questions and toxic comments, their accuracy was mostly around 70% on the Kaggle[13] website. So far with our Multi-LSTM ( ver_2 in this paper ) network(https://github.com/LanaChen0/Predict_News), the accuracy of news prediction can be up to 100%. Sadly, the corpora we used are from single-source media.

Future work may manage to collect multiple sources from different media and utilize the model on more complicated features. There are several examples, firstly, classify the author of the last forty chapters of Dream of the Red Chamber[14], which is still an unsolved mystery. Secondly, this model can be utilized to classify the types of websites that a customer is browsing, and consequently identify the target customers for the advertisers. Thirdly, be an assistant in the medical industry by identifying a specific disease from its symptoms. Last but not least, in reading comprehension, for instance, categorizing and predicting the specific era a historical article is referring to, such as Cradle of civilization, Bronze Age, or Iron Age. And with the utilization, it has its potential in library books sorting. We hope our work on predicting news can not only help simplify the pre-processing of fake news detection but also do precise news classification.

---

[13] https://www.kaggle.com/c/quora-insincere-questions-classification/overview

[14] https://en.wikipedia.org/wiki/Dream_of_the_Red_Chamber